%% file: emnlp2020.tex
\documentclass[11pt,a4paper]{article}
\usepackage[hyperref]{emnlp2020}
\usepackage{times}
\usepackage{latexsym}

\usepackage{microtype}

\usepackage{epsfig}
\usepackage{graphicx}
\usepackage{amssymb}
\usepackage{bm}

\usepackage{amsmath}
\usepackage{csquotes}
\usepackage{amsfonts}
\usepackage{multirow}
\usepackage{booktabs}

\aclfinalcopy 

\title{Vokenization: Improving Language Understanding with \\ Contextualized, Visual-Grounded Supervision}
 
\author{Hao Tan \;\;\;\;\;\;\; Mohit Bansal \\
  UNC Chapel Hill \\
  {\tt \{haotan, mbansal\}@cs.unc.edu} \\
 }

\date{}

\begin{document}
\maketitle
\begin{abstract}
Humans learn language by listening, speaking, writing, reading, and also, via interaction with the multimodal real world.
Existing language pre-training frameworks show the effectiveness of text-only self-supervision while we explore the idea of a visually-supervised language model in this paper.
We find that the main reason hindering this exploration is the large divergence in magnitude and distributions between the visually-grounded language datasets and pure-language corpora.
Therefore, we develop a technique named ``vokenization" that extrapolates multimodal alignments to language-only data by contextually mapping language tokens to their related images (which we call ``vokens").
The ``vokenizer'' is trained on relatively small image captioning datasets and
we then apply it to generate vokens for large language corpora.
Trained with these contextually generated vokens, our visually-supervised language models show consistent improvements over self-supervised alternatives on multiple pure-language tasks such as GLUE, SQuAD, and SWAG.\footnote{Code and pre-trained models publicly available at: \href{https://github.com/airsplay/vokenization}{https://github.com/airsplay/vokenization}.}

\end{abstract}

\input{1_intro}

\input{2_background}
\input{3_methods}

\input{4_results}

\input{5_analysis}
\input{6_related}

\input{7_conclusion}

\section*{Acknowledgement}
 We thank the reviewers and Yixin Nie and Jie Lei for their helpful discussions. This work was supported by ARO-YIP Award W911NF-18-1-0336, DARPA MCS Grant N66001-19-2-4031, a Google Focused Research Award, and a Bloomberg Data Science Ph.D. Fellowship. The views, opinions, and/or findings contained in this article are those of the authors and not of the funding agency.

\bibliographystyle{acl_natbib}
\bibliography{emnlp2020}

\appendix

\input{appendix}

\end{document}

%% file: 1_intro.tex
\section{Introduction}
\label{sec:intro}
Most humans learn language understanding from multiple modalities rather than only from the text and audio, especially using the visual modality.
As claimed in \newcite{bloom2002children}, visual pointing is an essential step for most children to learn meanings of words.
However, existing language pre-training frameworks are driven by contextual learning which only takes the language context as self-supervision.
For example, word2vec~\cite{mikolov2013distributed} takes surrounding bag-of-words;
ELMo~\cite{peters2018deep} and GPT~\cite{radford2018improving} take succeeding contexts;
and BERT~\cite{devlin2019bert} takes randomly masked tokens.
Although these self-supervised frameworks have achieved strong progress towards understanding human language, they did not borrow grounding information from the external visual world (see related motivations in recent work by \newcite{bender2020climbing} and \newcite{bisk2020experience}).

\begin{figure}[t]
\centering
\includegraphics[width=0.48\textwidth]{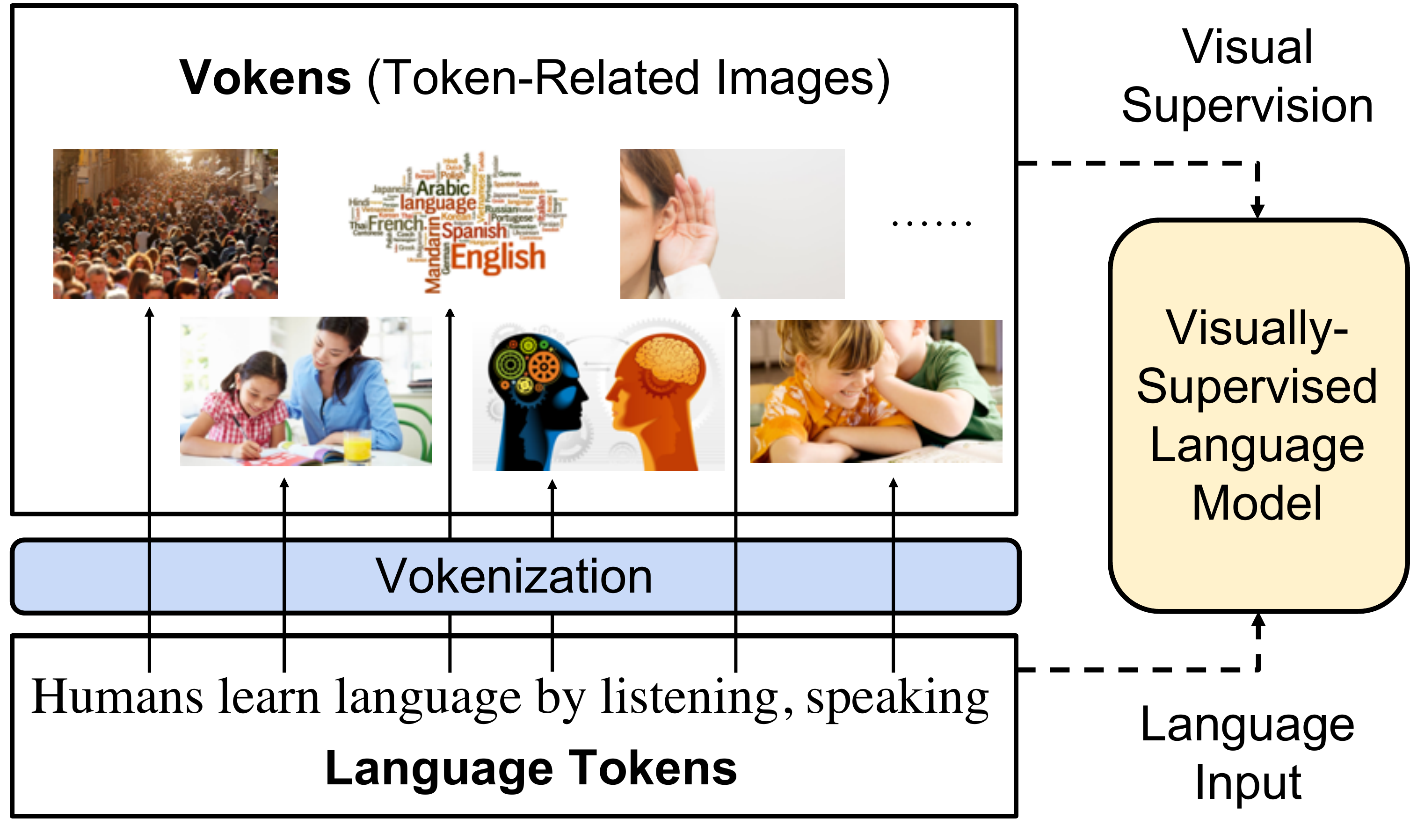}
\caption{
We visually supervise the language model with token-related images. We call these images vokens (visualized tokens) and develop a vokenization process to contextually generate them.
}
\vspace{-5pt}
\label{fig:task}
\end{figure}

In this paper, we introduce the visually-supervised language model that simulates human language learning with visual pointing~\cite{bloom2002children}. 
As shown in Fig.~\ref{fig:task}, this model takes language tokens as input and uses token-related images as visual supervision.
We name these images as \emph{vokens} (i.e., visualized tokens), since they act as visualizations of the corresponding tokens. 
Assuming that a large aligned token-voken dataset exists, the model could learn from these vokens via voken-prediction tasks.

\begin{figure*}[t]
\centering
\includegraphics[width=0.95\textwidth]{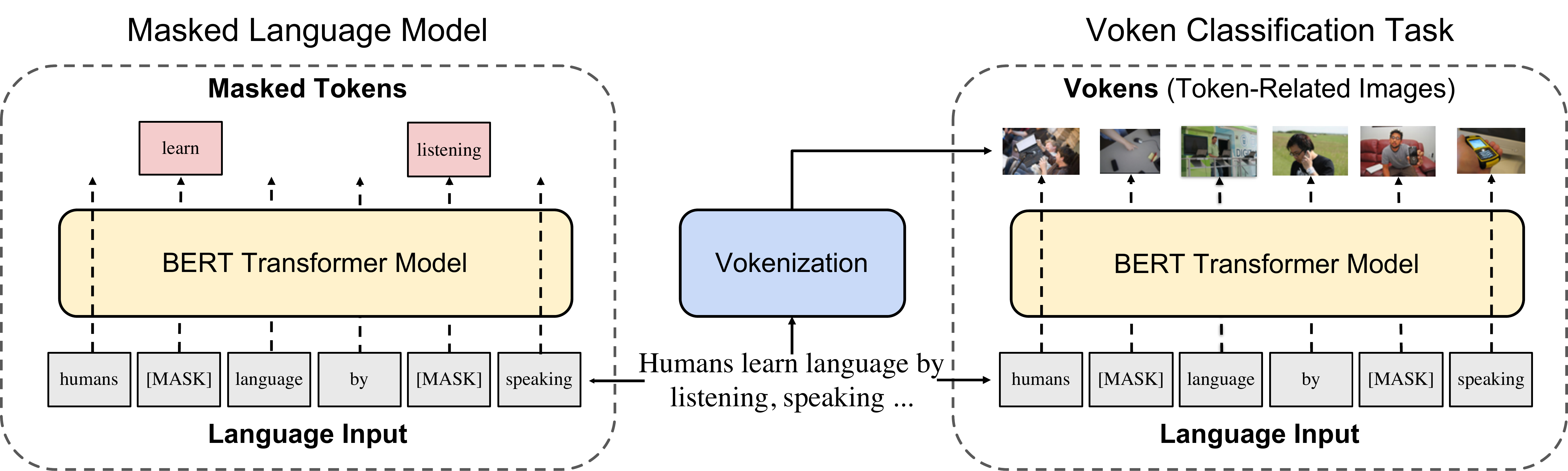}
\caption{
Illustration of the BERT transformer model trained with a visually-supervised language model with two objectives: masked language model (on the left) and voken classification (on the right).
The first objective (used in original BERT pre-training) predicts the masked tokens as self-supervision while the second objective predicts the corresponding vokens (contextually generated by our vokenization process) as external visual supervision.
Since the inputs are the same, we optimize the two objectives simultaneously and share the model weights.
}
\vspace{-10pt}
\label{fig:vslm}
\end{figure*}

Unfortunately, such an aligned token-voken dataset is currently unavailable and hence there are two main challenges in creating it from visually-grounded language datasets.
First, there is a large discrepancy between visually-grounded language (which provides innate visual grounding supervision) and other types of natural language.
For example, about 120M tokens are available in visually-grounded language datasets~\cite{tan2019lxmert, chen2019uniter}, which is far less compared to the 3,300M tokens in BERT training data and 220B tokens in T5~\cite{raffel2019exploring}.
Grounded language also prefers short and instructive descriptions, and thus has different distributions of sentence lengths and active words to other language types.
Second, most of the words in natural language are not visually grounded, hence this challenges the premise in creating visual supervision.
With an approximate estimation, the ratio of grounded tokens is only about $28\%$ in English Wikipedia.
This low grounding ratio leads to low coverage of visual supervision in previous approaches~\cite{frome2013devise, kiela2018learning}.

To resolve the above two challenges, we propose our \emph{vokenization} method (as shown in Fig.~\ref{fig:task}) that contextually maps the tokens to the visualized tokens (i.e., vokens) by retrieval.
Instead of directly supervising the language model with visually grounded language datasets (e.g., MS COCO~\cite{lin2014microsoft}), we use these relative small datasets to train the vokenization processor (i.e., the \emph{vokenizer}).
We then generate vokens for large language corpora (e.g., English Wikipedia), and our visually-supervised language model will take the input supervision from these large datasets, thus bridging the gap between different data sources, which solves the first challenge.
The second challenge of low grounding ratio seems to be an inherent characteristic of language; however, we observe that some non-visually-grounded tokens can be effectively mapped to related images when considering its context, e.g., the abstract word ``angry" in the sentence ``an angry cat lies on my leg".
This observation is realized by our \emph{contextual} token-image matching model (defined in Sec.~\ref{sec:model}) inside our vokenization processor, where we map tokens to images by viewing the sentence as the context.

Using our proposed vokenizer with a contextualized token-image matching model, we generate vokens for English Wikipedia.
Supervised by these generated vokens, we show consistent improvements upon a BERT model on several diverse NLP tasks such as GLUE~\cite{wang2018glue}, SQuAD~\cite{rajpurkar2016squad}, and SWAG~\cite{zellers2018swagaf}.
We also show the transferability of our vokens to other frameworks (i.e., RoBERTa).

%% file: 2_background.tex
\begin{table*}[]
\small
\begin{tabular}{@{}lccccccc@{}}
\toprule
\small Dataset       & \small \# of Tokens & \small \# of Sents & \small Vocab. Size & \small Tokens \#/ Sent. & \small 1-Gram JSD & \small 2-Gram JSD & \small Grounding Ratio \\ \midrule
MS COCO       & 7.0M        & 0.6M          & 9K       & 11.8                      &    0.15              &    0.27              & 54.8\%          \\
VG            & 29.2M       & 5.3M          & 13K      & 5.5                       &    0.16              &    0.28              & 57.6\%          \\
CC            & 29.9M       & 2.8M          & 17K      & 10.7                      &    0.09              &    0.20               & 41.7\%          \\ \midrule
Wiki103       & 111M        & 4.2M          & 29K      & 26.5                      &    0.01              &    0.05              & 26.6\%          \\
Eng Wiki      & 2889M       & 120M          & 29K      & 24.1                      &    0.00              &    0.00              & 27.7\%          \\ 
CNN/DM        & 294M        & 10.9M         & 28K      & 26.9                      &    0.04              &    0.10              & 28.3\%          \\ 
\bottomrule
\end{tabular}
\vspace{-5pt}
\caption{
Statistics of image-captioning dataset and other natural language corpora. 
VG, CC, Eng Wiki, and CNN/DM denote Visual Genome, Conceptual Captions, English Wikipedia, and CNN/Daily Mail, respectively.
JSD represents Jensen–Shannon divergence to the English Wikipedia corpus.
A large discrepancy exists between the visually grounded captioning and general language corpora.
}
\label{table:grounded_language_stat}
\vspace{-8pt}
\end{table*}

\section{Visually-Supervised Language Models}
Contextual language representation learning is driven by self-supervision without considering explicit connections (grounding) to the external world.
In this section, we illustrate the idea of a visually-supervised language model and discuss the challenges of creating its visual supervision.
\subsection{Vokens: Visualized Tokens}
\label{sec:voken}
To provide visual supervision to the language model, we assume a text corpus where each token is aligned with a related image (although these voken annotations currently do not exist, we will try to generate vokens next in Sec.~\ref{sec:vokenizer} by the vokenization process).
Hence, these images could be considered as visualizations of tokens and we name them as `vokens'.
Based on these vokens, we propose a new pre-training task for language: voken classification.

\subsection{The Voken-Classification Task}
\label{sec:voken_classification_task}
Most language backbone models (e.g., ELMo~\cite{peters2018deep}, GPT~\cite{radford2018improving}, BERT~\cite{devlin2019bert}) output a localized feature representation $\{\bm h_i\}$ for each token in a sentence $s=\{w_i\}$.
Thus it allows adding a token-level classification task without modifying the model architecture.
Suppose the vokens come from a finite set $\mathbb{X}$, we convert the hidden output $\bm h_i$ to a probability distribution $p_i$ with a linear layer and a softmax layer, then the voken classification loss is the negative log probability of all  corresponding vokens:
\begin{align*}
    \bm{h}_1, \bm{h}_2, \ldots, \bm{h}_l &= \mathrm{lm}(w_1, w_2, \ldots, w_l) \\
      p_i(v \mid s) &= \mathrm{softmax}_v \{ W \, \bm h_i + b\} \\
\mathcal{L}_\textsc{voken-cls}(s) &= - \sum_{i=1}^{l} \log p_i\left(v(w_i; s) \mid s \right)
\end{align*}
This task could be easily integrated into current language pre-training frameworks, and  
we next show an example.

\paragraph{Example: Visually-Supervised BERT} 
Fig.~\ref{fig:vslm} shows an example realization of the voken-classification task that provides visual supervision to BERT~\cite{devlin2019bert}.
The original BERT pre-training mainly relies on the task of masked language model\footnote{The next-sentence prediction task is removed in RoBERTa~\cite{liu2019roberta} and XLM~\cite{lample2019cross} and the fine-tuning results are not largely affected.} (illustrated on the left side of Fig.~\ref{fig:vslm}): tokens are randomly masked and the model needs to predict these missing tokens from language context.
For simplicity, we use $s$ and $\hat{s}$ to denote the set of tokens and masked tokens, separately. 
The unmasked tokens are the set difference $s \setminus \hat{s}$.
Suppose $q_i$ is the conditional probability distribution of the $i$-th token, the Masked Language Model (MLM) loss is the negative log-likelihood of the masked tokens:
\begin{align*}
\mathcal{L}_\textsc{mlm}(s, \hat{s}) = -\sum_{w_i \in \hat{s}} \log q_i\left(w_i \mid s \setminus \hat{s} \right)
\end{align*}
Without changing the model and model's inputs, we calculate the voken-classification loss for all tokens (illustrated on the right side of Fig.~\ref{fig:vslm}):
\begin{align*}
\mathcal{L}_\textsc{voken-cls}(s, \hat{s}) = -\sum_{w_i \in s} \log p_i\left(v(w_i;s) \mid s \setminus \hat{s} \right)
\end{align*}
The visually-supervised masked language model takes the sum of these two losses with a ratio $\lambda$.
\begin{align}
    \mathcal{L}_\textsc{vlm} (s, \hat{s}) = \mathcal{L}_\textsc{voken-cls}(s, \hat{s}) + \lambda \mathcal{L}_\textsc{mlm}(s, \hat{s}) 
\label{eqn:vlm}
\end{align}

\subsection{Two Challenges in Creating Vokens}
\label{sec:challenge}
Previous sections illustrate the potential external supervision by assuming the existence of vokens.
However, we are currently lacking the dense annotations from tokens to images. 
The most similar concept to vokens is phrase localization (e.g., in Flickr30K entities~\cite{flickr30k, flickrentitiesijcv}).
Because the process of collecting phrase localization is costly, the coverage and the amount of annotations cannot meet our requirements.\footnote{Recently, a concurrent work \newcite{pont2019connecting} releases localized narratives. The tokens are aligned with image pixels instead of images. 
}
Apart from phrase localization, the most promising data source is image captioning datasets with sentence-to-image mappings (or discovered from multimodal documents, as in \newcite{hessel2019unsupervised}).
Image captions belong to a specific type of language called \emph{grounded language}~\cite{roy2002learning, hermann2017grounded}, which has an explicit grounding to external existence or physical actions.
However, grounded language has a large discrepancy to other types of natural language  (e.g., News, Wiki, and Textbooks). 
To illustrate this, we list key statistics of three image-captioning dataset (i.e., MS COCO~\cite{lin2014microsoft}, Visual Genome~\cite{krishna2017visual}, and Conceptual Captions~\cite{sharma2018conceptual}) and three language corpora of other language types (i.e., Wiki103~\cite{merity2016pointer}, English Wiki, and CNN/Daily Mail~\cite{see2017get}) in Table~\ref{table:grounded_language_stat}.
This discrepancy between grounded language and other types of natural language leads to two challenges:

\vspace{2pt}
\noindent\textbf{A. Different Distributions between Grounded Language and Other Natural Language Corpora.} \
Sentences belonging to grounded language are usually short and informative, e.g., the average sentence length in MS COCO is $11.8$, which is much shorter than the average sentence length of $24.1$ in English Wiki.
The vocabulary\footnote{The vocabulary is calculated following \newcite{karpathy2015deep} where the words with $>5$ occurrence is counted.} of MS COCO only covers around one-third of token types~\cite{smith2019contextual} in English Wiki.
There is also a large divergence of the 1-Gram and 2-Gram distributions (measured by Jensen–Shannon divergence) between grounded language dataset and the English Wikipedia.
Lastly, the amount of tokens in grounded language corpora are also orders of magnitude smaller than commonly-used Wikipedia.

\vspace{2pt}
\noindent\textbf{B. Low Grounding Ratio in Natural Language.} \
The grounding ratio is defined as the percentage of visually grounded tokens in the dataset.
Visually grounded tokens (e.g., concrete nouns) are the token types that are naturally related to specific visual contents (e.g., `cat', `cake', `clock'). 
Since a precise list of such token types is hard to define, we thus estimate the grounding ratio based on existing grounded language corpora.
Specifically, we consider a token type with more than $100$ occurrences in MS COCO (after removing all stop words) as visually-grounded.
A sample of these token types could be found in the Appendix.
As shown in the last column of Table~\ref{table:grounded_language_stat}, the grounding ratio of English Wiki is $27.7\%$, which is almost half of that in Visual Genome.

To address these two challenges, we propose a vokenizer with contextual token-image matching models next in Sec.~\ref{sec:vokenizer}.

%% file: 3_methods.tex
\section{Vokenization}
\label{sec:vokenizer}
In the previous section, we discuss the potential of using vokens (i.e., visualized tokens) as visual supervision to the language model, and also demonstrate the large gap between currently available resources (i.e., annotated dataset) and the desired requirements.
Hence, in this section, we develop a framework that can generate vokens.
As shown in Fig.~\ref{fig:vslm}, the general idea is that we learn a ``vokenizer" from image-captioning dataset and use it to annotate large language corpora (i.e., English Wiki), thus bridging the gap between grounded language and other types of natural language.
We start by illustrating the vokenization process and then describe how we implement it.
\subsection{The Vokenization Process}
\label{sec:vokenization}
As shown in Fig.~\ref{fig:task} and Fig.~\ref{fig:vslm}, vokenization is the process to assign each token $w_i$ in a sentence $s=(w_1, w_2, \ldots, w_l)$ with a relevant image $v(w_i; s)$. 
We call this image $v(w_i; s)$ as a `voken' (visualized token).
Instead of creating this image with generative models, we retrieve an image from a set of images $\mathbb{X} = \{x_1, x_2, \ldots, x_n\}$ regarding a token-image-relevance scoring function $r_\theta(w_i, x; s)$.
This scoring function $r_\theta(w_i, x; s)$, parameterized by $\theta$, measures the relevance between the token $w_i$ in the sentence $s$ and the image $x$.
We here assume that the optimal parameter of this function is $\theta^*$ and will discuss the details of formulations later.
The voken $v(w_i; s)$ related to a token $w_i$ in the sentence $s$ is realized as the image $x \in \mathbb{X}$ that maximizes their relevance score $r_{\theta^*}$:
\begin{align*}
v(w_i; s) = \arg\max\nolimits_{x \in \mathbb{V}} \, r_{{\theta}^*}(w_i, x; s)
\end{align*}
Since the image set $\mathbb{X}$ indeed builds a finite vocabulary for vokens, we could utilize the voken-classification task (formulated in Sec.~\ref{sec:voken_classification_task}) to visually supervise the language model training.
We next talk about the detailed implementation of this vokenization process.

\subsection{Contextual Token-Image Matching Model}
\label{sec:model}
Lying in the core of the vokenization process is a contextual token-image matching model. 
The model takes a sentence $s$ and an image $x$ as input, and the sentence $s$ is composed of a sequence of tokens $\{w_1, w_2, \ldots, w_l\}$.  
The output $r_\theta(w_i, x;s)$ is the relevance score 
between the token $w_i \in s$ and the image $x$ while considering the whole sentence $s$ as a context.

\paragraph{Modeling} 
To model the relevance score function $r_\theta(w_i, x;s)$, we factorize it as an inner product of the language feature representation $\bm f_\theta(w_i; s)$ and the visual feature representation  $\bm{g}_\theta(x)$:
\begin{align*}
    r_\theta(w_i, x; s) = \bm f_\theta(w_i; s)^\intercal \bm g_\theta(x)
\end{align*}
These two feature representations 
are generated by language and visual encoders respectively.
The language encoder first uses a pre-trained $\text{BERT}_\text{BASE}$~\cite{devlin2019bert} model to contextually embed the discrete tokens $\{w_i\}$ into hidden-output vectors  $\{\bm{h}_i\}$:
\begin{align*}
    \bm{h}_1, \bm{h}_2, \ldots, \bm{h}_l = \mathit{bert}(w_1, w_2, \ldots, w_l)
\end{align*} 
Then we apply a multi-layer perceptron (MLP) $\mathit{w\mbox{\_}mlp}_\theta$ to down project the hidden output $\bm{h}_i$.
In order to simplify the retrieval process in Sec.~\ref{sec:vokenization}, the final language features are normalized to norm-1 vectors by dividing their Euclidean norms:
\begin{align*}
    \bm f_\theta(w_i; s) = \frac{\mathit{w\mbox{\_}mlp}_\theta(\bm{h}_i)}{\Vert \mathit{w\mbox{\_}mlp}_{\theta}(\bm{h}_i)  \Vert}
\end{align*}
On the other side, the visual encoder first extracts the visual embedding $e$ from a pre-trained ResNeXt~\cite{xie2017aggregated}.
Similar to the language encoder, an MLP layer $\mathit{x\mbox{\_}mlp}_\theta$ and an L2-normalization layer are applied subsequently:
\begin{align*}
    \bm{e} & = \mathrm{ResNeXt}(x) \\ 
    \bm g_\theta(x) & = \frac{\mathit{x\mbox{\_}mlp}_\theta(\bm{e})}{\Vert \mathit{x\mbox{\_}mlp}_\theta(\bm{e}) \Vert}
\end{align*}

\paragraph{Training} 
Since the dense annotations from tokens to images are lacking and hard to generate (illustrated in Sec.~\ref{sec:challenge}),
we thus alternatively 
train the token-image matching model
from weak supervision in image-captioning datasets (e.g., MS COCO~\cite{lin2014microsoft}).
These datasets are comprised of sentence-image pairs $\{(s_k, x_k)\}$ where the sentence $s_k$ describes the visual content in image $x_k$.
To build alignments between tokens and images, we pair all tokens in a sentence $s_k$ with the image $x_k$.
The model is then optimized by 
maximizing the relevance score of these aligned token-image pairs over unaligned pairs.

Without loss of generality, assuming $(s, x)$ is an image-captioning data point,
we randomly sample another image $x'$ with the condition $x' \neq x$.
We then use hinge loss to optimize the weight $\theta$ so that the score of the positive token-image pair $r_\theta(w_i, x;s)$ aims to be larger than the negative pair $r_\theta(w_i, x';s)$ by at least a margin $M$.
\vspace{-5pt}
\begin{align*}
    \vspace{-5pt}
    \mathcal{L}_\theta(s, x, x') = \sum_{i=1}^{l} \max\{0, M & - r_\theta(w_i, x;s) \\ 
    & + r_\theta(w_i, x';s)\}
\end{align*}
Intuitively, minimizing this hinge loss $\max\{0,$ $M-\mathit{pos} + \mathit{neg}\}$ will try to increase the score of the positive pair and decrease the score of the negative pair when the score difference is smaller than the margin $M$.
Otherwise (if the difference is $\geq$ margin $M$), the two scores remain unchanged.

\begin{figure}[t]
\centering
\includegraphics[width=0.49\textwidth]{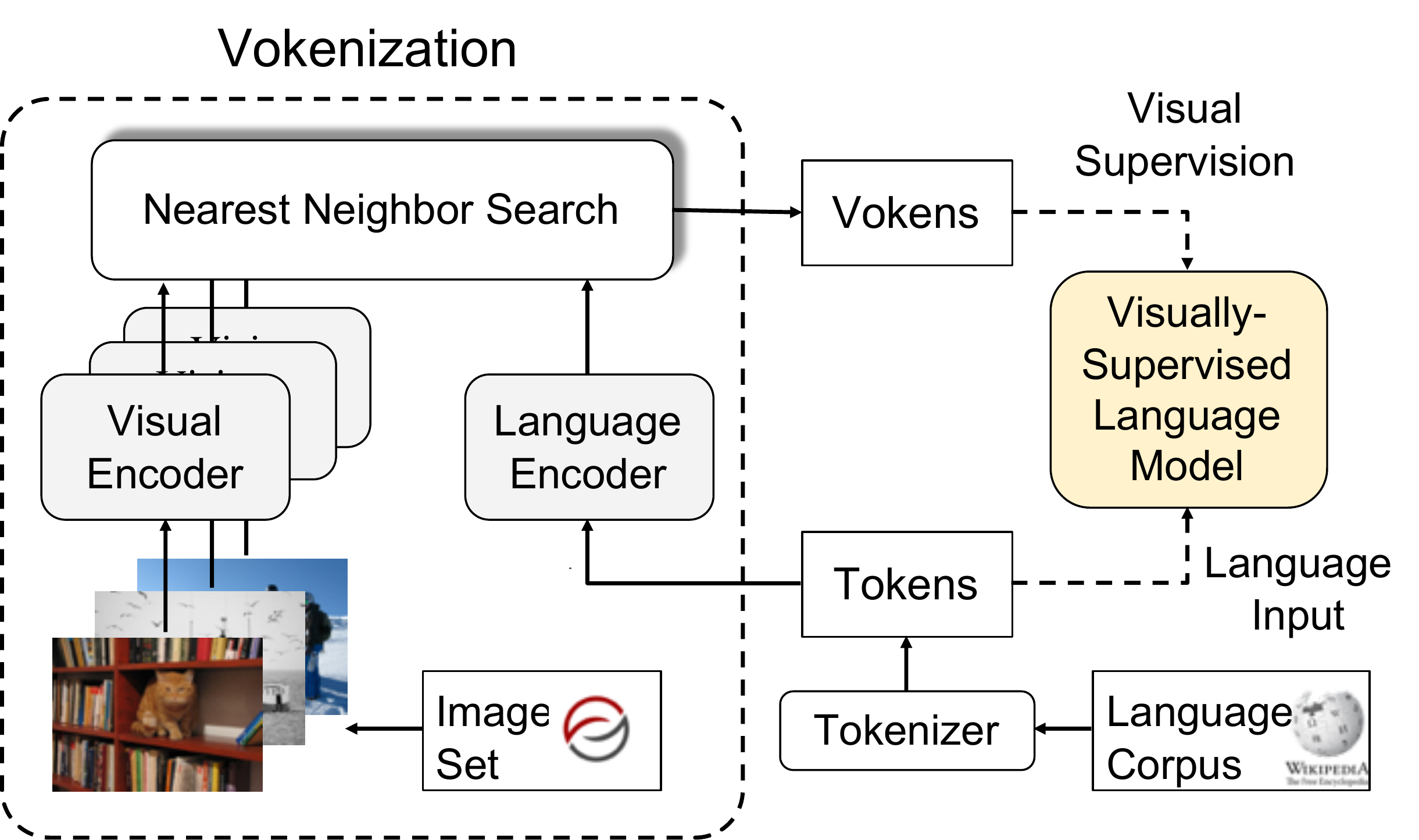}
\vspace{-16pt}
\caption{
Implementation of our vokenization process.
For the tokens in language corpora, we contextually retrieved images (with nearest neighbor search) from the image set as vokens.
These generated vokens are then used as the visual supervision to the language model.
}
\label{fig:vokenization}
\vspace{-10pt}
\end{figure}

\paragraph{Inference}
Given that the relevance score is factorized as the inner product of feature representations $\bm f_\theta(w_i;s)$ and $\bm g_\theta(v)$,
the retrieval problem in Sec.~\ref{sec:vokenization} could be formulated as Maximum Inner Product Search~\cite{mussmann2016learning}).
Moreover, since the vectors are norm-1, the vector with the maximum inner product is identical to the closest vector in the Euclidean space (i.e., Nearest Neighbor~\cite{knuth1973art}). 
We illustrate the detailed implementation in Fig.~\ref{fig:vokenization}.

\subsection{Revokenization}
\label{sec:revokenization}
A constraint of the vokenization process in Sec.~\ref{sec:vokenization} is that the vokens depend on the actual tokenizer of the language encoder in Sec.~\ref{sec:model}.
Since different frameworks utilize a various range of tokenizers, this constraint limits the transferability of vokens between different frameworks.
Instead of binding our vokenizer to a specific pre-training framework (e.g., BERT), we want to enable its extensibility to other frameworks (e.g., RoBERTa).
Thus, we introduce a ``revokenization" technique to address this limitation.

Given two different tokenizers $T_1$ and $T_2$, 
they tokenize a sentence $s$ into two different sequences of tokens:
$T_1(s) = \left(w_1, w_2, \ldots, w_l\right)$
and $T_2(s) = \left(u_1, u_2, \ldots, u_m\right)$.
Without loss of generality, assuming the vokenizer is built based on the first tokenizer $T_1$, the standard vokenization process will generate a sequence of vokens $\{v(w_i;s)\}_{i=1}^l$ which are one-to-one aligned with the tokens $\{w_i\}_{i=1}^l$.
Our goal is to transfer these $w$-related vokens to the $u$-related vokens generated by $T_2$.
We adapt the idea of ``nearest neighbor algorithm"~\cite{altman1992introduction}  here. 
For a given token $u_j$, among all $w$'s, we select the one that overlaps the most with $u_j$ and record it as $w_{\mathit{ind}(j)}$. 
The voken for $u_j$ is defined as the  voken for its ``nearest neighbor" $w_{\mathit{ind}(j)}$:
\vspace{-5pt}
\begin{align*}
\vspace{-5pt}
&v(u_j; s) := v(w_{\mathit{ind}(j)}; s) \\ 
\mathit{ind}(j) &= \arg\max\nolimits_{i=1}^l \mathrm{overlap}(w_i, u_j) 
\vspace{-5pt}
\end{align*}
The overlapping of two tokens are further quantified by the intersection-over-union (i.e., Jaccard index, defined as $\mathrm{IoU}(A, B)\mbox{=}\frac{\vert A\cap B\vert}{\vert A \cup B\vert}$) of 
their ranges
in the raw sentence $s$.

%% file: 4_results.tex
\begin{table*}[]
\centering
\resizebox{0.9\textwidth}{!}{%
\begin{tabular}{@{}lcccccccc@{}}
\toprule
Method                 & \small SST-2& \small QNLI & \small QQP     & \small MNLI & \small SQuAD v1.1 & \small SQuAD v2.0 & \small SWAG & \small Avg.\\ \midrule
$\text{BERT}_\text{6L/512H}$ 
                      & 88.0& 85.2& 87.1   & 77.9&71.3/80.2 &57.2/60.8 & 56.2 & 75.6\\
$\text{BERT}_\text{6L/512H}$  + Voken-cls
                      & 89.7& 85.0& 87.3   & 78.6&71.5/80.2 &61.3/64.6 & 58.2 & 76.8\\
$\text{BERT}_\text{12L/768H}$ 
                       & 89.3& 87.9& 83.2   & 79.4&77.0/85.3 &67.7/71.1 & 65.7 & 79.4\\
$\text{BERT}_\text{12L/768H}$  + Voken-cls
                       & \textbf{92.2}& \textbf{88.6}& \textbf{88.6}   & \textbf{82.6}&\textbf{78.8/86.7} &68.1/71.2 &\textbf{70.6} & \textbf{82.1}\\
\midrule                                                                                      
$\text{RoBERTa}_\text{\,6L/512H}$ 
                      & 87.8& 82.4& 85.2   & 73.1&50.9/61.9 &49.6/52.7 & 55.1 & 70.2\\
$\text{RoBERTa}_\text{\,6L/512H}$  + Voken-cls
                      & 87.8& 85.1& 85.3   & 76.5&55.0/66.4 &50.9/54.1 & 60.0 & 72.6\\ 
$\text{RoBERTa}_\text{\,12L/768H}$ 
                       & 89.2& 87.5& 86.2   & 79.0&70.2/79.9 &59.2/63.1 & 65.2& 77.6 \\
$\text{RoBERTa}_\text{\,12L/768H}$  + Voken-cls
                       & \textbf{90.5}& \textbf{89.2}& \textbf{87.8}   & \textbf{81.0}&\textbf{73.0/82.5} &\textbf{65.9/69.3} & \textbf{70.4} & \textbf{80.6}\\ \bottomrule
\end{tabular}}
\vspace{-4pt}
\caption{
Fine-tuning results of different pre-trained models w/ or w/o the voken classification task (denoted as ``Voken-cls").
SQuAD results are ``exact match"/``F1".
The results which significantly outperform the second-best ones are marked in bold.
The averages of metrics (denoted as ``Avg.") show improvement from voken supervisions.
}
\label{table:results}
\vspace{-5pt}
\end{table*}
\section{Experimental Setups and Results}

\subsection{Pre-training Data and Fine-tuning Tasks}
We train our model on English Wikipedia~\footnote{BERT~\cite{devlin2019bert} also uses Toronto Books Corpus~\cite{zhu2015aligning}.
However, the dataset is not publicly released. We thus exclude it in our study to ensure reproducibility.} 
and its featured subset Wiki103~\cite{merity2016pointer}.
We use our vokenizer to generate vokens for these two datasets as well.
The pre-trained models are then fine-tuned on GLUE~\cite{wang2018glue}, SQuAD~\cite{rajpurkar2016squad,rajpurkar2018know}, and SWAG~\cite{zellers2018swagaf} to assess the pre-training performance.
Since some smaller tasks in GLUE are reported as unstable~\cite{dodge2020fine}, recent papers (e.g., \newcite{li2020train}) only report on selected tasks.
We follow this trend and evaluate on the four largest datasets (i.e., SST-2~\cite{socher2013recursive}, QNLI~\cite{rajpurkar2016squad}, QQP~\cite{iyer2017first}, MNLI~\cite{williams2018broad}).\footnote{The size of the used four dataset range from $60$K to $400$ while the omitted dataset range from $0.6$K to $8.5$K.}.

\subsection{Implementation Details}
\label{sec:impl_details}
We train our contextual token-image matching model (in Sec.~\ref{sec:model}) on MS COCO image captioning dataset for $20$ epochs.
The concatenation of the last 4 layers of BERT outputs and $\mathrm{ResNeXt\mbox{-}101\mbox{-}32x8d}$ features are used as language hidden states and visual embedding, respectively.
Both multi-layer perceptrons $\mathit{w\mbox{\_}mlp}_\theta$ and $\mathit{x\mbox{\_}mlp}_\theta$ have two fully-connected layers with $256$-dimensional intermediate outputs (followed by ReLU activation) and $64$-dimensional final outputs.
The two backbone models BERT~\cite{devlin2019bert} and ResNeXt~\cite{xie2017aggregated} are not fine-tuned.
We set the hinge loss margin $M$ to $0.5$.
During the vokenization process of English Wikipedia and Wiki103, we use the faiss~\cite{johnson2019billion} library to speed up the nearest neighbor search.
The vokens are retrieved from the Visual Genome images that are not used in MS COCO.
We fix a voken size of $50000$.

When pre-training the model on pure language corpus, we unify the training protocols to avoid possible side effects.
We follow previous works to conduct two simplifications: 1. Removing the next-sentence-prediction task~\cite{liu2019roberta} 2. Using fixed sequence length~\cite{conneau2019unsupervised} of $128$.
We take the $12$-layer $\text{BERT}_\text{BASE}$ model of $768$ hidden dimensions and train it on English Wikipedia for $200$K steps from scratch.
We also take a reduced $6$-layer model and train it on Wiki103 for $40$ epochs ($160$K steps) because this reduced model could not fit the full English Wikipedia dataset.

Since we only use the vokens in the supervision, the voken-classification task does not bring additional parameters to the language model but needs more computations.
We thus adjust the training steps for pure masked-language-model (MLM) training accordingly for a fair comparison.
The loss ratio $\lambda\mbox{=}1.0$ in Eqn.~\ref{eqn:vlm} is not tuned because of limited budget.
All pre-training processes take batch sizes of $256$ and learning rates of $2e\mbox{-}4$. 
For fine-tuning tasks, we report the results on the validation sets. 
We train $3$ epochs with a learning rate of $1e\mbox{-}4$ and a batch-size of $32$ for all tasks in GLUE.
The hyper-parameters for SQuAD, SWAG are borrowed from BERT.

\begin{table*}[]
\centering
\resizebox{0.9\textwidth}{!}{%
\begin{tabular}{@{}lcccccc@{}}
\toprule
\small Model                                             & \small{Init. with BERT?}  & \small{Diff. to BERT Weight}& \small SST-2 & \small QNLI & \small QQP  & \small MNLI    \\ \midrule
ViLBERT\small~\cite{lu2019vilbert}                      & Yes                       & 0.0e-3                      &90.3  &89.6  &88.4  &82.4     \\
VL-BERT\small~\cite{su2019vl}                           & Yes                       & 6.4e-3                      &90.1  &89.5  &88.6  &82.9     \\
VisualBERT\small~\cite{li2019visualbert}                & Yes                       & 6.5e-3                      &90.3  &88.9  &88.4  &82.4     \\
Oscar\small~\cite{li2020oscar}                          & Yes                       & 41.6e-3                     &87.3  &50.5  &86.6  &77.3     \\  
LXMERT~\small\cite{tan2019lxmert}                 & No                        & 42.0e-3                     &82.4  &50.5  &79.8  &31.8     \\ \midrule
$\text{BERT}_\text{BASE}$\small~\cite{devlin2019bert}   & -                         & 0.0e-3                      &90.3  &89.6  &88.4  &82.4     \\
$\text{BERT}_\text{BASE}$ + Weight Noise          &-                          & 6.5e-3                      &89.9  &89.9  &88.4  &82.3   \\
\bottomrule
\end{tabular}
}
\caption{
Results of vision-and-language pre-trained models on GLUE tasks.
We also provide BERT models w/ and w/o weight noise as baselines.
}
\vspace{-10pt}
\label{table:vlp}
\end{table*}

\begin{table}[]
\centering
\resizebox{0.40\textwidth}{!}{%
\begin{tabular}{@{}lcccc@{}}
\toprule
  Pre-trained on         & \small SST-2 & \small QNLI & \small QQP  & \small MNLI    \\ \midrule
 MS COCO         &83.7  &60.6  &82.1  &69.3     \\
 Wiki103*        &85.8  &77.9  &84.8  &73.9     \\ 
No Pre-train     &77.1  &50.5  &31.6  &31.8     \\
\bottomrule
\end{tabular}
}
\caption{
Results of BERT models pre-trained on captions in MS COCO and a reduced version of  Wiki103 dataset (denoted as Wiki103*). 
Models without pre-training are taken as a baseline.
}
\vspace{-10pt}
\label{table:mscoco}
\end{table}

\subsection{Results}
\label{sec:results}
As reported in Table~\ref{table:results}, we fine-tune the pre-trained models on different natural-language tasks.
The models are either pre-trained with masked language model (e.g., ``$\text{BERT}_\text{6L/512H}$'') or pre-trained with masked language model with an additional voken-classification task (e.g., ``$\text{BERT}_\text{6L/512H}$+Voken-cls'') 
following Eqn.~\ref{eqn:vlm}.
The default metric is accuracy.
Following \newcite{wang2018glue}, we report the average of F1 and accuracy for QQP. 
For SQuAD, we report the exact matching and F1 score respectively.
We also compute macro-averages for evaluated tasks (denoted as ``Avg.'' in the last column) as a general indicator. 
Although the different architectures of models (i.e., 6L/512H and 12L/768H) affect the fine-tuning results,
the voken-classification task consistently improves the downstream tasks' performance and achieves large average gains.
We also show the transferability of our vokenizer to the RoBERTa model and observe the same phenomenon as that in BERT.

%% file: 5_analysis.tex
\begin{table*}[]
\centering
\begin{tabular}{@{}lllcccc@{}}
\toprule
Method              & Retrieval     & Supervision     &   SST-2 & QNLI & QQP  & MNLI    \\ \midrule
SentLabel           & Sent-level   & Sent-level      &88.3  &86.1  &86.9  &78.0    \\
Propagated          & Sent-level   & Token-level      &88.9  &87.9  &88.1  &80.2     \\
Term Frequency      & Token-level   & Token-level      &89.0  &86.9  &85.5  &79.8     \\ 
\midrule
Vokens              & Contextual Token-level   & Token-level     &92.2  &88.6  &88.6  &82.6     \\
\bottomrule
\end{tabular}
\vspace{-4pt}
\caption{
Comparisons of sentence-level (denoted as ``Sent-level") and token-level approaches. 
Token-level approaches outperform the sentence-level approaches from both retrieval-method and supervision perspective.
}
\label{table:sent_vs_token}
\vspace{-10pt}
\end{table*}

\section{Analysis}
\label{sec:ablation}

\subsection{Limit of Visually-Grounded Language}
\label{sec:vlp_limit}
In Sec.~\ref{sec:challenge}, we illustrated the differences between (visually-)grounded-language datasets and other natural-language corpora by demonstrating their contrasting statistics.
In this section, 
we study the models trained with grounded language and show their ineffectiveness on pure-language tasks.
We first investigate vision-and-language pre-training frameworks, which succeed on multimodal tasks.
As shown in Table~\ref{table:vlp}, when fine-tuning them on pure-language tasks, the results are generally lower than the pre-trained BERT model.\footnote{ViLBERT~\cite{lu2019vilbert} freezes the BERT weight in its training thus their results are the same to BERT; Uniter~\cite{chen2019uniter} shrinks its vocab thus is not shown.}
Although these frameworks are different in multiple ways, 
the only remarkable factor to the fine-tuning results is the BERT-weight initialization.
Moreover, we also show that these models are similar to a BERT model with a random weight noise of the same magnitude.
We thus claim that vision-and-language pre-training on visually-grounded language dataset currently might not help the pure-language tasks.
Note that the BERT results in Table~\ref{table:results} are not fairly comparable to the results in Table~\ref{table:vlp} because the original BERT model~\cite{devlin2019bert} also uses Toronto Books Corpus~\cite{zhu2015aligning}. 
Unfortunately, this dataset is not publicly available and hence we exclude it.
According to \newcite{raffel2019exploring}, the exclusion of Toronto Books Corpus downgrades the results and we observe the same tendency here (comparing $\text{BERT}_\text{12L/768H}$ in Table~\ref{table:results} and $\text{BERT}_\text{BASE}$ in Table~\ref{table:vlp}). 

Besides these existing models, 
we next investigate the BERT models trained with masked language model on grounded language data (i.e., MS COCO).
A control experiment is built by shrinking the Wiki103 to the same token amount as MS COCO.
We also provide the BERT model trained from scratch as a baseline.
As shown in Table~\ref{table:mscoco}, the model trained with MS COCO is significantly worse than the model trained with Wiki103 on all downstream tasks.
The reason might be the large discrepancy between visually-grounded language and other types of language as shown in Sec.~\ref{sec:challenge}.

\subsection{Token-Level vs. Sentence-Level Approaches}
\label{sec:token_vs_sent}
In Sec.~\ref{sec:intro}, we stated the drawbacks of the purely sentence-level and token-level approaches, then introduce the contextual token-level approach (i.e., the contextual token-image matching model in Sec.~\ref{sec:model}) which combines these two approaches.
In this section, we demonstrate a careful comparison between our vokenization process and the other two approaches from two perspectives: the retrieval methods and the supervision types.
Experiments are conducted with the same hyper-parameters and dataset as ``$\text{BERT}_\text{12L/768H}$+Voken-cls'' in Table~\ref{table:results}.
\paragraph{Sentence-Level Retrieval}
To conduct sentence-level retrieval, we first adapt the contextual token-image matching model in Sec.~\ref{sec:model} to a sentence-image matching model (details in Appendix).
We then retrieve a related image for each sentence.
As shown in Table~\ref{table:sent_vs_token}, these retrieved images are used as two kinds of supervisions by putting classifiers at different places:
in the row ``SentLabel", we provide sentence-level supervision by using the classifier to predict the label for the whole sentence (similar to the BERT's ``next-sentence prediction" (NSP) task);
and in the row ``Propagated", we provide token-level supervision by propagating sentence-level labels to all tokens in the sentences, and apply the classifier at each token (similar to our voken-classification task).
The results of both kinds of supervisions are lower than our proposed vokens (in the row ``Vokens").
One possible reason for these lower results is that finding an image that conveys the meaning of the whole sentence is hard.
We also find that dense token-level supervision also outperforms the sentence-level supervision.

\paragraph{Token-level Retrieval}
Our proposed vokenization process is viewed as contextual token-level retrieval, which grounds tokens with whole sentences as context.
We here consider a purely token-level retrieval method regarding term frequencies.
The term frequency $\mathit{tf}(\mathit{tok}, x_i)$~\cite{manning2008introduction} is calculated based on the occurrence $\#(\mathit{tok}, x_i)$ of the token $\mathit{tok}$ in the image $x_i$'s captions.
\begin{align*}
\vspace{-5pt}
    \mathit{tf}(\mathit{tok}, x_i)  = \frac{\#(\mathit{tok}, x_i)}{\sum_{\mathit{tok}'}\#(\mathit{tok}', x_i)} 
\end{align*}
We then convert this term frequency to the conditional distribution via Boltzmann distribution:
\begin{align*}
    p(x_i \mid \mathit{tok}) = \frac{\exp{\left(\mathit{tf}(\mathit{tok}, x_i) / \gamma \right)}}{\sum_{x'} \exp{\left(\mathit{tf}(\mathit{tok}, x') / \gamma\right)}}
\end{align*}
where $\gamma$ is temperature. 
We stochastically map the tokens to images with this conditional distribution $p(x_i \mid \mathit{tok})$.
The results trained with these special vokens are shown in Table~\ref{table:sent_vs_token} as ``Term Frequency".
Overall, token-level supervision is still better than the sentence-level supervision (as in the row ``SentLabel").
However, among the models trained with token-level supervision, this token-level retrieval method neglects the contextual information thus is worse compared with sentence-level (in the row ``Propagated") and contextual token-level retrieval methods (in the row ``Voken") .

\begin{figure}[t]
\centering
\includegraphics[width=0.47\textwidth]{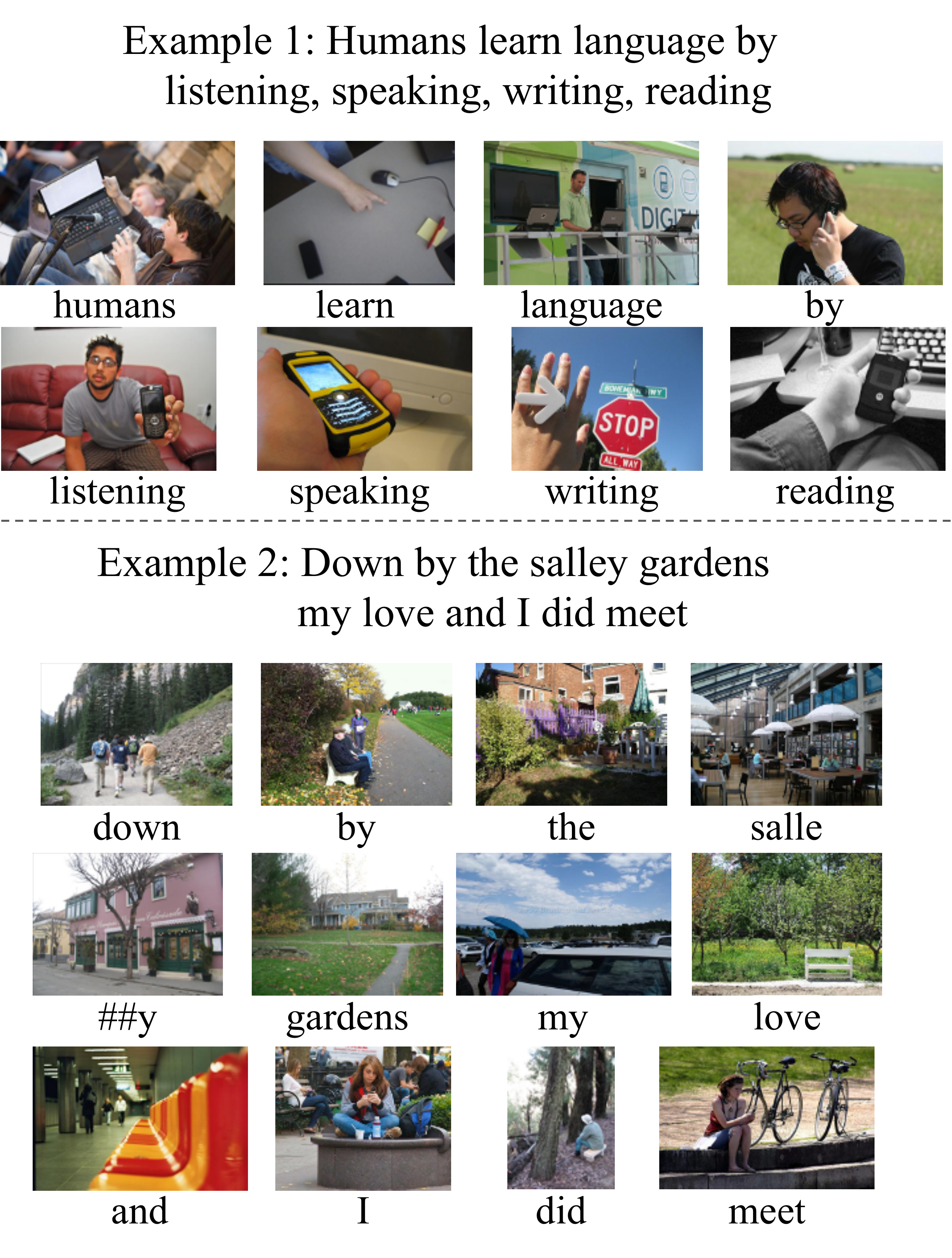}
\vspace{-10pt}
\caption{
Visualization of model-generated vokens. Example 1 takes the leading sentence of this paper while Examples 2 takes Yeats's poet.
}
\vspace{-10pt}
\label{fig:examples}
\end{figure}

\subsection{Visualization of Vokens}
In Fig.~\ref{fig:examples}, we visualize our generated vokens.
The first example takes the leading sentence in our paper (without commas), which is also used in the imaginary example in Fig.~\ref{fig:task}.
We also vokenize another sentence from William Yeats's poet ``Down by the Salley Gardens'' in Fig.~\ref{fig:examples}.
Although the vokenizer is trained on image-captioning datasets without localizing token-to-image annotations,
the vokenizer shows a strong selectivity: different images are selected w.r.t the tokens.
The contextual token-level retrieval could also disambiguate certain tokens (e.g., ``down" in Example 2) with the help of its context.
When the \emph{unique} related image is hard to define, our vokenizer aims to ground the non-concrete tokens (e.g., ``by"/``and"/``the") to relevant images: the voken for the token ``by'' in Example 2 (of Fig.~\ref{fig:examples}) is better aligned with the [centering token, context] pair than the voken for the same token ``by'' in Example 1. 
This related visual information helps understand the language and leads to the improvement in Table \ref{table:results}. 
On the other hand, some tokens are not faithfully grounded (e.g., ``writing" in Example 1) and we also observe a shift in alignment (e.g., the relevant image for the phrase ``my love" in Example 2 is aligned to ``my" instead of ``love").
These misalignments are possibly caused by the limitations of sentence-image weak supervision in our training data since the strong token-image annotations are not available.

%% file: 6_related.tex
\section{Related Work}
\paragraph{Language (Model) Pre-training} 
Language pre-training has moved from token-level pre-training~\cite{mikolov2013distributed, pennington2014glove} to sentence-level pre-training~\cite{le2014distributed, kiros2015skip, conneau2017supervised, dai2015semi}.
Recently, a set of works~\cite{peters2018deep, radford2018improving, devlin2019bert, yang2019xlnet, liu2019roberta, clark2019electra, lan2019albert} bring back token-level supervision with contextual language encoders (e.g., based on an LSTM~\cite{hochreiter1997long} and Transformers~\cite{vaswani2017attention}).
This tendency inspires the design of our vokenizer in merging previous sentence-level~\cite{frome2013devise} and token-level~\cite{kiela2018learning} approaches into a contextual token-level approach.

\paragraph{Vision-and-Language Pre-training} 
Since language models are trained with self-supervision without knowing the connection to the visual world, 
vision-and-language pre-training~\cite{li2019visualbert, lu2019vilbert, tan2019lxmert, chen2019uniter, su2019vl, zhou2019unified}  aims to build joint cross-modal representations and focuses on vision-and-language tasks.
Due to particularity of grounded language, these models are not able to improve pure language tasks as shown in Sec.~\ref{sec:vlp_limit}.

\paragraph{Visually-Aided Language Learning} 
Previous works use visual information to improve specific language tasks such as coreference resolution~\cite{kong2014you}, machine translation~\cite{elliott2016multi30k, ive2019distilling, wu2019predicting, Zhang2020Neural}, semantic parsing~\cite{christie2016resolving, shi2019visually, kojima2020learned}, and  bilingual lexicon learning~\cite{kiela2015visual, vulic2016multi}.
Our work has a focus on building a visually-supervised language pre-training frameworks to improve general language understanding.
Similar to our work, \newcite{frome2013devise, lazaridou2015combining, collell2017imagined, kiela2018learning, bordes2020incorporating} aim to improve language representation with visual information; 
however, most of these works focus on grounded language and hence might suffer from the large discrepancy that we discuss in Sec.~\ref{sec:challenge}.

%% file: 7_conclusion.tex
\section{Conclusion}
In this paper, we explored the possibility of utilizing visual supervision to language encoders.
In order to overcome the challenges in grounded language, we develop the vokenizer with contextual token-image matching models and use it to vokenize the language corpus.
Supervised by these generated vokens, we observe a significant improvement over the purely self-supervised language model on multiple language tasks.

%% file: appendix.tex
\section{Appendices}

\subsection{Full Implementation Details}
We train our contextual token-image matching model (in Sec. 3.1) on MS COCO image captioning dataset\footnote{\url{http://cocodataset.org/}} for $20$ epochs.
The concatenation of the last 4 layers of BERT outputs (following \newcite{devlin2019bert}) and mean pooling of $\mathrm{ResNeXt\mbox{-}101\mbox{-}32x8d}$ feature maps are used as features for tokens and the images. 
For both multi-layer perceptrons $\mathit{w\mbox{\_}mlp}_\theta$ and $\mathit{x\mbox{\_}mlp}_\theta$, we use two fully-connected layers with ReLU activation, where the output dimensions of the two layers are $256$ and $64$, accordingly.
We only train the modules marked with $\theta$, i.e., the two backbone models BERT~\cite{devlin2019bert} and ResNeXt~\cite{xie2017aggregated} are not fine-tuned.
Since we normalize the features $g(w_i;s)$ and $f(v)$ to be norm-1 vectors, the relevance score thus takes the range from $[-1, 1]$ (from the Cauchy Inequality). 
The margin $M$ in hinge loss is set to $0.5$.

During the vokenization process, we use the faiss~\cite{johnson2019billion} library to speed up the nearest neighbor search.
The vokenization runs at a speed of 100K tokens / second with 4 Titan V100 GPU.
Thus the vokenization of the full Wikipedia is finished in 8 hours.
When transferring vokens to other pre-training frameworks, 
revokenization does not need the GPU computation and runs as fast as the tokenization.
The vokens are retrieved from the Visual Genome images which are not used in MS COCO (our training dataset).
We take a voken size of $50000$.

When pre-training the model on pure language corpus, we unify the training process to avoid possible side effects from different training protocols.
We follow previous work to conduct two simplifications: 1. Removing the next-sentence-prediction task~\cite{liu2019roberta} 2. Using fixed sequence length~\cite{conneau2019unsupervised} of $128$.
We take the $12$-layer $\text{BERT}_\text{BASE}$ model of $768$ hidden dimensions and train it on English Wikipedia\footnote{Downloaded with \url{https://github.com/attardi/wikiextractor}} for $200$K steps from scratch.
We also take a reduced $6$-layer model and train it on Wiki103\footnote{https://www.salesforce.com/products/einstein/ai-research/the-wikitext-dependency-language-modeling-dataset/} for $40$ epochs ($160$K steps) from scratch because this reduced model does not fit well on the full Wikipedia dataset.
The voken classification task will not bring additional parameters to the language encoder (with $110$M parameters) but need more computations, we thus adjust the training steps for pure masked-language-model (MLM) training for a fair comparison.
It results in around $10$\% more training steps in pure MLM training.
All models take batch sizes of $256$ and a learning rate of $2e\mbox{-}4$.

For fine-tuning tasks, instead of high-cost hyper-parameter sweeping in BERT~\cite{devlin2019bert}, we train $3$ epochs with a learning rate of $1e\mbox{-}4$ and a batch-size of $32$ for all tasks in GLUE.
The hyper-parameters for SQuAD and SWAG are borrowed from the BERT paper~\cite{devlin2019bert}.
On SQuAD v1.1, we fine-tune  for $3$ epochs with a learning rate of $5e\mbox{-}5$ and a batch size of $32$.
On SQuAD v2.0, we fine-tune  for $2$ epochs with a learning rate of $5e\mbox{-}5$ and a batch size of $48$.
On SWAG, we fine-tune  for $3$ epochs with a learning rate of $2e\mbox{-}5$ and a batch size of $16$. 

The whole framework is built on PyTorch~\cite{paszke2019pytorch}. 
The implementations of BERT~\cite{devlin2019bert} and RoBERTa~\cite{liu2019roberta} are borrowed from PyTorch Transformers~\cite{Wolf2019HuggingFacesTS}\footnote{ \url{https://github.com/huggingface/transformers}}.
All evaluation code is from the PyTorch Transformers as well.

\subsection{Visually Grounded Token Types}
In Sec.2.3, we estimate the visually grounded token types with the help of MS COCO~\cite{lin2014microsoft} dataset.
We here randomly sample a list of the $2406$ grounded tokens used in the estimation:

photograph, tv, skyscraper, \#\#bery, wooded, little, stands, away, storage, mound, pouring, rail, \#\#fl, eye, \#\#ke, flown, skiing, plate, movie, dead, tossing, couple, racing, dust, licking, palm, stroll, granite, bananas, ledge, chained, monument, individuals, part, exhibit, softball, second, bow, ones, shop, beverages, sandy, sink, angle, \#\#ia, gives, music, leading, carrying, cookies, reading, faced, \#\#k, kid, \#\#ged, playing, winds, saddle, stunts, squat, cabinets, rusty, matching, biker, let, standing, pan, smiles, train, sky, passing, woman, military, feeder, lot, hydra, party, \#\#l, furnished, rides, strip, \#\#field, tin, crouched, courtyard, nicely, screens, us, lie, waving, process, equipment, structure, fore, barrier, \#\#li, beside, toast, catching, tracks

\subsection{Maximum Inner Product Search of Norm-1 Vectors}
In Sec. 3.1, we normalize the vector to norm-1 vectors thus the Maximum Inner Product Search~\cite{mussmann2016learning} is equivalent to Nearest Neighbor~\cite{knuth1973art}.
Here, we give a simple proof.
Suppose $\bm x$ and $\bm y$ are two vectors of the same dimension,  we have
\begin{align}
\Vert \bm x - \bm y \Vert^2 & = \Vert \bm x \Vert^2 + \Vert \bm y \Vert^2 - 2 \bm x^\intercal \bm y\\
& = 2 - 2 
\bm x^\intercal \bm y 
\end{align}
Without loss of generality, we assume that there is a unique vector $\bm \hat{y} \in \mathbb{Y}$ with the maximum inner product and thus
\begin{align}
    \bm \hat{y} = \arg\min_x \Vert \bm x - \bm y \Vert = \arg\max_x \bm x^\intercal \bm y
\end{align}

\subsection{Details of Sentence-level Retrieval in Analysis}
In Sec. 3.1, we consider a contextual token-image matching model with relevance score $r_\theta(w, x; s)$.
To do sentence-level retrieval, we modify it into a sentence-image matching score $r'_\theta(x, s)$, and trained it with:
\begin{align*}
    \tilde{\mathcal{L}}_\theta(s, x, x') =  \max\{0, M & - r'_\theta(x, s) \\
    & + r'_\theta(x', s)\}
\end{align*}
The score is also factorized as the dot product of the visual representation and the language representation.
However, the language representation here is the sentence embedding (the output for the first token \text{CLS}).

We retrieve the image from the same image set $\mathbb{V}$ as vokenization and with the similar Maximum Inner Product Search method:
\begin{align*}
v(s) = \arg\max\nolimits_{x \in \mathbb{V}} \, r'_{\theta^*}(x, s)
\end{align*}
These retrieved images as used as the label for the whole sentence.

\begin{table}[]
\centering
\begin{tabular}{@{}lcccccc@{}}
\toprule
Voken Type         & SST-2  & QNLI  & QQP     & MNLI  \\ 
\midrule
\multicolumn{5}{@{}l@{}}{\small \textbf{Alternative Choices}} \\
Random             & 89.1 & 87.6 & 86.6    & 80.0 \\
Shuffle            & 89.2 & 87.3 & 86.1    & 80.2 \\
Tokens             & 89.7 & 88.8 & 87.2    & 80.8 \\
\midrule
\multicolumn{5}{@{}l@{}}{\small \textbf{Reference Models}} \\
Voken Only         & 89.8 & 87.8 & 86.2    & 81.7  \\ 
No Voken           & 89.3 & 87.9 & 83.2    & 79.4  \\ 
Voken              & 92.2 & 88.6 & 88.6    & 82.6  \\ 
\bottomrule
\end{tabular}
\caption{Results of different strategies that replace the standard vokenization process. }
\label{table:alternatives}
\end{table}

\subsection{Details of Token-level Retrieval in Analysis}
In the purely token-level retrieval, we consider the image-captioning sentences as documents and uses traditional IR methods to index them.
In order to increase the size of `documents',   we aggregate the data from VQA~\cite{antol2015vqa} and Visual Genome~\cite{krishna2017visual}, besides the existing MS COCO~\cite{lin2014microsoft} dataset.
We also find that the temperature $\gamma\mbox{=}0.01$ gives a reasonable retrieval distribution and use it in our experiment.

\subsection{Voken Ablation Studies}
In Table~\ref{table:alternatives}, we show several approaches that provide alternative voken-like labels to our model.
\paragraph{Random} We replace the vokens with random int from $\{1 \ldots \Vert \mathbb{V} \Vert \}$, where $\mathbb{V}$ is the ``vocabulary'' of all vokens.
\paragraph{Shuffle} 
In order to prove that the order of vokens would affect the results, we shuffle the vokens in each batch and use it as supervision.
\paragraph{Tokens}
We here directly use the original tokens in replace of the vokens to see whether any dense supervision could improve the model.

As shown in Table~\ref{table:alternatives}, all these results are lower than the reference vokenization strategy. 

\subsection{Correlations between Improvements and Grounding Ratio}
In order to understand where the improvements in the performance are coming from, we also study the correlation between the improvement in results and the visual grounding ratio (approximately measured in the same way as Sec.~\ref{sec:challenge}).
We found that the datasets with higher grounding ratio (e.g., MNLI~\cite{williams2018broad}) get significant improvements while the datasets (e.g., QNLI~\cite{rajpurkar2016squad}) with relatively lower grounding ratio do not benefit much from the visual supervision.
The dataset MNLI is built from multiple genre (the original SNLI dataset is in fact built from the Flickr images thus has a strong  visual connection) and QNLI is purely based on English Wikipedia (The same as SQuAD~\cite{rajpurkar2016squad}). 
These correlations may indicate that the visual supervision helps build a better understanding of visually grounded tokens.
Although we used contextual information to map non-grounded words to related images through vokenization, the effectiveness of this mapping relies on the original grounding ratio of the data.

%% file: emnlp2020.bbl
\begin{thebibliography}{75}
\expandafter\ifx\csname natexlab\endcsname\relax\def\natexlab#1{#1}\fi

\bibitem[{Altman(1992)}]{altman1992introduction}
NS~Altman. 1992.
\newblock An introduction to kernel and nearest-neighbor nonparametric
  regression.
\newblock \emph{The American statistician}, 46(3):175--184.

\bibitem[{Antol et~al.(2015)Antol, Agrawal, Lu, Mitchell, Batra,
  Lawrence~Zitnick, and Parikh}]{antol2015vqa}
Stanislaw Antol, Aishwarya Agrawal, Jiasen Lu, Margaret Mitchell, Dhruv Batra,
  C~Lawrence~Zitnick, and Devi Parikh. 2015.
\newblock Vqa: Visual question answering.
\newblock In \emph{Proceedings of the IEEE international conference on computer
  vision}, pages 2425--2433.

\bibitem[{Bender and Koller(2020)}]{bender2020climbing}
Emily~M Bender and Alexander Koller. 2020.
\newblock Climbing towards nlu: On meaning, form, and understanding in the age
  of data.
\newblock In \emph{ACL}.

\bibitem[{Bisk et~al.(2020)Bisk, Holtzman, Thomason, Andreas, Bengio, Chai,
  Lapata, Lazaridou, May, Nisnevich, Pinto, and Turian}]{bisk2020experience}
Yonatan Bisk, Ari Holtzman, Jesse Thomason, Jacob Andreas, Yoshua Bengio, Joyce
  Chai, Mirella Lapata, Angeliki Lazaridou, Jonathan May, Aleksandr Nisnevich,
  Nicolas Pinto, and Joseph Turian. 2020.
\newblock Experience grounds language.
\newblock In \emph{EMNLP}.

\bibitem[{Bloom(2002)}]{bloom2002children}
Paul Bloom. 2002.
\newblock \emph{How children learn the meanings of words}.
\newblock MIT press.

\bibitem[{Bordes et~al.(2019)Bordes, Zablocki, Soulier, Piwowarski, and
  Gallinari}]{bordes2020incorporating}
Patrick Bordes, {\'E}loi Zablocki, Laure Soulier, Benjamin Piwowarski, and
  Patrick Gallinari. 2019.
\newblock Incorporating visual semantics into sentence representations within a
  grounded space.
\newblock In \emph{EMNLP}.

\bibitem[{Chen et~al.(2019)Chen, Li, Yu, Kholy, Ahmed, Gan, Cheng, and
  Liu}]{chen2019uniter}
Yen-Chun Chen, Linjie Li, Licheng Yu, Ahmed~El Kholy, Faisal Ahmed, Zhe Gan,
  Yu~Cheng, and Jingjing Liu. 2019.
\newblock Uniter: Learning universal image-text representations.
\newblock \emph{arXiv preprint arXiv:1909.11740}.

\bibitem[{Christie et~al.(2016)Christie, Laddha, Agrawal, Antol, Goyal,
  Kochersberger, and Batra}]{christie2016resolving}
Gordon Christie, Ankit Laddha, Aishwarya Agrawal, Stanislaw Antol, Yash Goyal,
  Kevin Kochersberger, and Dhruv Batra. 2016.
\newblock Resolving language and vision ambiguities together: Joint
  segmentation \& prepositional attachment resolution in captioned scenes.
\newblock In \emph{Proceedings of the 2016 Conference on Empirical Methods in
  Natural Language Processing}, pages 1493--1503.

\bibitem[{Clark et~al.(2019)Clark, Luong, Le, and Manning}]{clark2019electra}
Kevin Clark, Minh-Thang Luong, Quoc~V Le, and Christopher~D Manning. 2019.
\newblock Electra: Pre-training text encoders as discriminators rather than
  generators.
\newblock In \emph{International Conference on Learning Representations}.

\bibitem[{Collell et~al.(2017)Collell, Zhang, and Moens}]{collell2017imagined}
Guillem Collell, Ted Zhang, and Marie-Francine Moens. 2017.
\newblock Imagined visual representations as multimodal embeddings.
\newblock In \emph{Thirty-First AAAI Conference on Artificial Intelligence}.

\bibitem[{Conneau et~al.(2020)Conneau, Khandelwal, Goyal, Chaudhary, Wenzek,
  Guzm{\'a}n, Grave, Ott, Zettlemoyer, and Stoyanov}]{conneau2019unsupervised}
Alexis Conneau, Kartikay Khandelwal, Naman Goyal, Vishrav Chaudhary, Guillaume
  Wenzek, Francisco Guzm{\'a}n, Edouard Grave, Myle Ott, Luke Zettlemoyer, and
  Veselin Stoyanov. 2020.
\newblock Unsupervised cross-lingual representation learning at scale.
\newblock In \emph{ACL}.

\bibitem[{Conneau et~al.(2017)Conneau, Kiela, Schwenk, Barrault, and
  Bordes}]{conneau2017supervised}
Alexis Conneau, Douwe Kiela, Holger Schwenk, Lo{\"\i}c Barrault, and Antoine
  Bordes. 2017.
\newblock Supervised learning of universal sentence representations from
  natural language inference data.
\newblock In \emph{Proceedings of the 2017 Conference on Empirical Methods in
  Natural Language Processing}, pages 670--680.

\bibitem[{Dai and Le(2015)}]{dai2015semi}
Andrew~M Dai and Quoc~V Le. 2015.
\newblock Semi-supervised sequence learning.
\newblock In \emph{Advances in neural information processing systems}, pages
  3079--3087.

\bibitem[{Devlin et~al.(2019)Devlin, Chang, Lee, and
  Toutanova}]{devlin2019bert}
Jacob Devlin, Ming-Wei Chang, Kenton Lee, and Kristina Toutanova. 2019.
\newblock Bert: Pre-training of deep bidirectional transformers for language
  understanding.
\newblock In \emph{Proceedings of the 2019 Conference of the North American
  Chapter of the Association for Computational Linguistics: Human Language
  Technologies, Volume 1 (Long and Short Papers)}, pages 4171--4186.

\bibitem[{Dodge et~al.(2020)Dodge, Ilharco, Schwartz, Farhadi, Hajishirzi, and
  Smith}]{dodge2020fine}
Jesse Dodge, Gabriel Ilharco, Roy Schwartz, Ali Farhadi, Hannaneh Hajishirzi,
  and Noah Smith. 2020.
\newblock Fine-tuning pretrained language models: Weight initializations, data
  orders, and early stopping.
\newblock \emph{arXiv preprint arXiv:2002.06305}.

\bibitem[{Elliott et~al.(2016)Elliott, Frank, Sima’an, and
  Specia}]{elliott2016multi30k}
Desmond Elliott, Stella Frank, Khalil Sima’an, and Lucia Specia. 2016.
\newblock Multi30k: Multilingual english-german image descriptions.
\newblock In \emph{Proceedings of the 5th Workshop on Vision and Language},
  pages 70--74.

\bibitem[{Frome et~al.(2013)Frome, Corrado, Shlens, Bengio, Dean, Ranzato, and
  Mikolov}]{frome2013devise}
Andrea Frome, Greg~S Corrado, Jon Shlens, Samy Bengio, Jeff Dean, Marc'Aurelio
  Ranzato, and Tomas Mikolov. 2013.
\newblock Devise: A deep visual-semantic embedding model.
\newblock In \emph{Advances in neural information processing systems}, pages
  2121--2129.

\bibitem[{Hermann et~al.(2017)Hermann, Hill, Green, Wang, Faulkner, Soyer,
  Szepesvari, Czarnecki, Jaderberg, Teplyashin et~al.}]{hermann2017grounded}
Karl~Moritz Hermann, Felix Hill, Simon Green, Fumin Wang, Ryan Faulkner, Hubert
  Soyer, David Szepesvari, Wojciech~Marian Czarnecki, Max Jaderberg, Denis
  Teplyashin, et~al. 2017.
\newblock Grounded language learning in a simulated 3d world.
\newblock \emph{arXiv preprint arXiv:1706.06551}.

\bibitem[{Hessel et~al.(2019)Hessel, Lee, and Mimno}]{hessel2019unsupervised}
Jack Hessel, Lillian Lee, and David Mimno. 2019.
\newblock Unsupervised discovery of multimodal links in multi-image,
  multi-sentence documents.
\newblock In \emph{EMNLP}.

\bibitem[{Hochreiter and Schmidhuber(1997)}]{hochreiter1997long}
Sepp Hochreiter and J{\"u}rgen Schmidhuber. 1997.
\newblock Long short-term memory.
\newblock \emph{Neural computation}, 9(8):1735--1780.

\bibitem[{Ive et~al.(2019)Ive, Madhyastha, and Specia}]{ive2019distilling}
Julia Ive, Pranava Madhyastha, and Lucia Specia. 2019.
\newblock Distilling translations with visual awareness.
\newblock In \emph{ACL}.

\bibitem[{Iyer et~al.(2017)Iyer, Dandekar, and Csernai}]{iyer2017first}
Shankar Iyer, Nikhil Dandekar, and Korn{\'e}l Csernai. 2017.
\newblock First quora dataset release: Question pairs.
\newblock \emph{data. quora. com}.

\bibitem[{Johnson et~al.(2019)Johnson, Douze, and
  J{\'e}gou}]{johnson2019billion}
Jeff Johnson, Matthijs Douze, and Herv{\'e} J{\'e}gou. 2019.
\newblock Billion-scale similarity search with gpus.
\newblock \emph{IEEE Transactions on Big Data}.

\bibitem[{Karpathy and Fei-Fei(2015)}]{karpathy2015deep}
Andrej Karpathy and Li~Fei-Fei. 2015.
\newblock Deep visual-semantic alignments for generating image descriptions.
\newblock In \emph{Proceedings of the IEEE conference on computer vision and
  pattern recognition}, pages 3128--3137.

\bibitem[{Kiela et~al.(2018)Kiela, Conneau, Jabri, and
  Nickel}]{kiela2018learning}
Douwe Kiela, Alexis Conneau, Allan Jabri, and Maximilian Nickel. 2018.
\newblock Learning visually grounded sentence representations.
\newblock In \emph{Proceedings of the 2018 Conference of the North American
  Chapter of the Association for Computational Linguistics: Human Language
  Technologies, Volume 1 (Long Papers)}, pages 408--418.

\bibitem[{Kiela et~al.(2015)Kiela, Vulic, and Clark}]{kiela2015visual}
Douwe Kiela, Ivan Vulic, and Stephen Clark. 2015.
\newblock Visual bilingual lexicon induction with transferred convnet features.
\newblock In \emph{Proceedings of the 2015 Conference on Empirical Methods in
  Natural Language Processing (EMNLP 2015)}. ACL; East Stroudsburg, PA.

\bibitem[{Kiros et~al.(2015)Kiros, Zhu, Salakhutdinov, Zemel, Urtasun,
  Torralba, and Fidler}]{kiros2015skip}
Ryan Kiros, Yukun Zhu, Russ~R Salakhutdinov, Richard Zemel, Raquel Urtasun,
  Antonio Torralba, and Sanja Fidler. 2015.
\newblock Skip-thought vectors.
\newblock In \emph{Advances in neural information processing systems}, pages
  3294--3302.

\bibitem[{Knuth(1973)}]{knuth1973art}
Donald~E Knuth. 1973.
\newblock The art of computer programming, volume 3: Searching and sorting.
\newblock \emph{Addison-Westley Publishing Company: Reading, MA}.

\bibitem[{Kojima et~al.(2020)Kojima, Averbuch-Elor, Rush, and
  Artzi}]{kojima2020learned}
Noriyuki Kojima, Hadar Averbuch-Elor, Alexander~M Rush, and Yoav Artzi. 2020.
\newblock What is learned in visually grounded neural syntax acquisition.
\newblock In \emph{ACL}.

\bibitem[{Kong et~al.(2014)Kong, Lin, Bansal, Urtasun, and
  Fidler}]{kong2014you}
Chen Kong, Dahua Lin, Mohit Bansal, Raquel Urtasun, and Sanja Fidler. 2014.
\newblock What are you talking about? text-to-image coreference.
\newblock In \emph{Proceedings of the IEEE conference on computer vision and
  pattern recognition}, pages 3558--3565.

\bibitem[{Krishna et~al.(2017)Krishna, Zhu, Groth, Johnson, Hata, Kravitz,
  Chen, Kalantidis, Li, Shamma et~al.}]{krishna2017visual}
Ranjay Krishna, Yuke Zhu, Oliver Groth, Justin Johnson, Kenji Hata, Joshua
  Kravitz, Stephanie Chen, Yannis Kalantidis, Li-Jia Li, David~A Shamma, et~al.
  2017.
\newblock Visual genome: Connecting language and vision using crowdsourced
  dense image annotations.
\newblock \emph{International Journal of Computer Vision}, 123(1):32--73.

\bibitem[{Lample and Conneau(2019)}]{lample2019cross}
Guillaume Lample and Alexis Conneau. 2019.
\newblock Cross-lingual language model pretraining.
\newblock \emph{Advances in Neural Information Processing Systems (NeurIPS)}.

\bibitem[{Lan et~al.(2019)Lan, Chen, Goodman, Gimpel, Sharma, and
  Soricut}]{lan2019albert}
Zhenzhong Lan, Mingda Chen, Sebastian Goodman, Kevin Gimpel, Piyush Sharma, and
  Radu Soricut. 2019.
\newblock Albert: A lite bert for self-supervised learning of language
  representations.
\newblock In \emph{International Conference on Learning Representations}.

\bibitem[{Lazaridou et~al.(2015)Lazaridou, Baroni
  et~al.}]{lazaridou2015combining}
Angeliki Lazaridou, Marco Baroni, et~al. 2015.
\newblock Combining language and vision with a multimodal skip-gram model.
\newblock In \emph{Proceedings of the 2015 Conference of the North American
  Chapter of the Association for Computational Linguistics: Human Language
  Technologies}, pages 153--163.

\bibitem[{Le and Mikolov(2014)}]{le2014distributed}
Quoc Le and Tomas Mikolov. 2014.
\newblock Distributed representations of sentences and documents.
\newblock In \emph{International conference on machine learning}, pages
  1188--1196.

\bibitem[{Li et~al.(2019)Li, Yatskar, Yin, Hsieh, and Chang}]{li2019visualbert}
Liunian~Harold Li, Mark Yatskar, Da~Yin, Cho-Jui Hsieh, and Kai-Wei Chang.
  2019.
\newblock Visualbert: A simple and performant baseline for vision and language.
\newblock \emph{arXiv preprint arXiv:1908.03557}.

\bibitem[{Li et~al.(2020{\natexlab{a}})Li, Yin, Li, Hu, Zhang, Zhang, Wang, Hu,
  Dong, Wei et~al.}]{li2020oscar}
Xiujun Li, Xi~Yin, Chunyuan Li, Xiaowei Hu, Pengchuan Zhang, Lei Zhang, Lijuan
  Wang, Houdong Hu, Li~Dong, Furu Wei, et~al. 2020{\natexlab{a}}.
\newblock Oscar: Object-semantics aligned pre-training for vision-language
  tasks.
\newblock \emph{arXiv preprint arXiv:2004.06165}.

\bibitem[{Li et~al.(2020{\natexlab{b}})Li, Wallace, Shen, Lin, Keutzer, Klein,
  and Gonzalez}]{li2020train}
Zhuohan Li, Eric Wallace, Sheng Shen, Kevin Lin, Kurt Keutzer, Dan Klein, and
  Joseph~E Gonzalez. 2020{\natexlab{b}}.
\newblock Train large, then compress: Rethinking model size for efficient
  training and inference of transformers.
\newblock In \emph{ICML}.

\bibitem[{Lin et~al.(2014)Lin, Maire, Belongie, Hays, Perona, Ramanan,
  Doll{\'a}r, and Zitnick}]{lin2014microsoft}
Tsung-Yi Lin, Michael Maire, Serge Belongie, James Hays, Pietro Perona, Deva
  Ramanan, Piotr Doll{\'a}r, and C~Lawrence Zitnick. 2014.
\newblock Microsoft coco: Common objects in context.
\newblock In \emph{European conference on computer vision}, pages 740--755.
  Springer.

\bibitem[{Liu et~al.(2019)Liu, Ott, Goyal, Du, Joshi, Chen, Levy, Lewis,
  Zettlemoyer, and Stoyanov}]{liu2019roberta}
Yinhan Liu, Myle Ott, Naman Goyal, Jingfei Du, Mandar Joshi, Danqi Chen, Omer
  Levy, Mike Lewis, Luke Zettlemoyer, and Veselin Stoyanov. 2019.
\newblock Roberta: A robustly optimized bert pretraining approach.
\newblock \emph{arXiv preprint arXiv:1907.11692}.

\bibitem[{Lu et~al.(2019)Lu, Batra, Parikh, and Lee}]{lu2019vilbert}
Jiasen Lu, Dhruv Batra, Devi Parikh, and Stefan Lee. 2019.
\newblock Vilbert: Pretraining task-agnostic visiolinguistic representations
  for vision-and-language tasks.
\newblock In \emph{Advances in Neural Information Processing Systems}, pages
  13--23.

\bibitem[{Manning et~al.(2008)Manning, Raghavan, and
  Sch{\"u}tze}]{manning2008introduction}
Christopher~D Manning, Prabhakar Raghavan, and Hinrich Sch{\"u}tze. 2008.
\newblock \emph{Introduction to information retrieval}.
\newblock Cambridge university press.

\bibitem[{Merity et~al.(2017)Merity, Xiong, Bradbury, and
  Socher}]{merity2016pointer}
Stephen Merity, Caiming Xiong, James Bradbury, and Richard Socher. 2017.
\newblock Pointer sentinel mixture models.
\newblock In \emph{ICLR}.

\bibitem[{Mikolov et~al.(2013)Mikolov, Sutskever, Chen, Corrado, and
  Dean}]{mikolov2013distributed}
Tomas Mikolov, Ilya Sutskever, Kai Chen, Greg~S Corrado, and Jeff Dean. 2013.
\newblock Distributed representations of words and phrases and their
  compositionality.
\newblock In \emph{Advances in neural information processing systems}, pages
  3111--3119.

\bibitem[{Mussmann and Ermon(2016)}]{mussmann2016learning}
Stephen Mussmann and Stefano Ermon. 2016.
\newblock Learning and inference via maximum inner product search.
\newblock In \emph{International Conference on Machine Learning}, pages
  2587--2596.

\bibitem[{Paszke et~al.(2019)Paszke, Gross, Massa, Lerer, Bradbury, Chanan,
  Killeen, Lin, Gimelshein, Antiga et~al.}]{paszke2019pytorch}
Adam Paszke, Sam Gross, Francisco Massa, Adam Lerer, James Bradbury, Gregory
  Chanan, Trevor Killeen, Zeming Lin, Natalia Gimelshein, Luca Antiga, et~al.
  2019.
\newblock Pytorch: An imperative style, high-performance deep learning library.
\newblock In \emph{Advances in Neural Information Processing Systems}, pages
  8024--8035.

\bibitem[{Pennington et~al.(2014)Pennington, Socher, and
  Manning}]{pennington2014glove}
Jeffrey Pennington, Richard Socher, and Christopher~D Manning. 2014.
\newblock Glove: Global vectors for word representation.
\newblock In \emph{Proceedings of the 2014 conference on empirical methods in
  natural language processing (EMNLP)}, pages 1532--1543.

\bibitem[{Peters et~al.(2018)Peters, Neumann, Iyyer, Gardner, Clark, Lee, and
  Zettlemoyer}]{peters2018deep}
Matthew Peters, Mark Neumann, Mohit Iyyer, Matt Gardner, Christopher Clark,
  Kenton Lee, and Luke Zettlemoyer. 2018.
\newblock Deep contextualized word representations.
\newblock In \emph{Proceedings of the 2018 Conference of the North American
  Chapter of the Association for Computational Linguistics: Human Language
  Technologies, Volume 1 (Long Papers)}, pages 2227--2237.

\bibitem[{Plummer et~al.(2017)Plummer, Wang, Cervantes, Caicedo, Hockenmaier,
  and Lazebnik}]{flickrentitiesijcv}
Bryan~A. Plummer, Liwei Wang, Christopher~M. Cervantes, Juan~C. Caicedo, Julia
  Hockenmaier, and Svetlana Lazebnik. 2017.
\newblock Flickr30k entities: Collecting region-to-phrase correspondences for
  richer image-to-sentence models.
\newblock \emph{IJCV}, 123(1):74--93.

\bibitem[{Pont-Tuset et~al.(2019)Pont-Tuset, Uijlings, Changpinyo, Soricut, and
  Ferrari}]{pont2019connecting}
Jordi Pont-Tuset, Jasper Uijlings, Soravit Changpinyo, Radu Soricut, and
  Vittorio Ferrari. 2019.
\newblock Connecting vision and language with localized narratives.
\newblock \emph{arXiv preprint arXiv:1912.03098}.

\bibitem[{Radford et~al.(2018)Radford, Narasimhan, Salimans, and
  Sutskever}]{radford2018improving}
Alec Radford, Karthik Narasimhan, Tim Salimans, and Ilya Sutskever. 2018.
\newblock Improving language understanding by generative pre-training.
\newblock \emph{URL https://s3-us-west-2. amazonaws.
  com/openai-assets/researchcovers/languageunsupervised/language understanding
  paper. pdf}.

\bibitem[{Raffel et~al.(2019)Raffel, Shazeer, Roberts, Lee, Narang, Matena,
  Zhou, Li, and Liu}]{raffel2019exploring}
Colin Raffel, Noam Shazeer, Adam Roberts, Katherine Lee, Sharan Narang, Michael
  Matena, Yanqi Zhou, Wei Li, and Peter~J Liu. 2019.
\newblock Exploring the limits of transfer learning with a unified text-to-text
  transformer.
\newblock \emph{arXiv preprint arXiv:1910.10683}.

\bibitem[{Rajpurkar et~al.(2018)Rajpurkar, Jia, and Liang}]{rajpurkar2018know}
Pranav Rajpurkar, Robin Jia, and Percy Liang. 2018.
\newblock Know what you don’t know: Unanswerable questions for squad.
\newblock In \emph{Proceedings of the 56th Annual Meeting of the Association
  for Computational Linguistics (Volume 2: Short Papers)}, pages 784--789.

\bibitem[{Rajpurkar et~al.(2016)Rajpurkar, Zhang, Lopyrev, and
  Liang}]{rajpurkar2016squad}
Pranav Rajpurkar, Jian Zhang, Konstantin Lopyrev, and Percy Liang. 2016.
\newblock Squad: 100,000+ questions for machine comprehension of text.
\newblock In \emph{Proceedings of the 2016 Conference on Empirical Methods in
  Natural Language Processing}, pages 2383--2392.

\bibitem[{Roy and Pentland(2002)}]{roy2002learning}
Deb~K Roy and Alex~P Pentland. 2002.
\newblock Learning words from sights and sounds: A computational model.
\newblock \emph{Cognitive science}, 26(1):113--146.

\bibitem[{See et~al.(2017)See, Liu, and Manning}]{see2017get}
Abigail See, Peter~J Liu, and Christopher~D Manning. 2017.
\newblock Get to the point: Summarization with pointer-generator networks.
\newblock In \emph{ACL}.

\bibitem[{Sharma et~al.(2018)Sharma, Ding, Goodman, and
  Soricut}]{sharma2018conceptual}
Piyush Sharma, Nan Ding, Sebastian Goodman, and Radu Soricut. 2018.
\newblock Conceptual captions: A cleaned, hypernymed, image alt-text dataset
  for automatic image captioning.
\newblock In \emph{Proceedings of ACL}.

\bibitem[{Shi et~al.(2019)Shi, Mao, Gimpel, and Livescu}]{shi2019visually}
Haoyue Shi, Jiayuan Mao, Kevin Gimpel, and Karen Livescu. 2019.
\newblock Visually grounded neural syntax acquisition.
\newblock In \emph{Proceedings of the 57th Annual Meeting of the Association
  for Computational Linguistics}.

\bibitem[{Smith(2019)}]{smith2019contextual}
Noah~A Smith. 2019.
\newblock Contextual word representations: A contextual introduction.
\newblock \emph{arXiv preprint arXiv:1902.06006}.

\bibitem[{Socher et~al.(2013)Socher, Perelygin, Wu, Chuang, Manning, Ng, and
  Potts}]{socher2013recursive}
Richard Socher, Alex Perelygin, Jean Wu, Jason Chuang, Christopher~D Manning,
  Andrew~Y Ng, and Christopher Potts. 2013.
\newblock Recursive deep models for semantic compositionality over a sentiment
  treebank.
\newblock In \emph{Proceedings of the 2013 conference on empirical methods in
  natural language processing}, pages 1631--1642.

\bibitem[{Su et~al.(2020)Su, Zhu, Cao, Li, Lu, Wei, and Dai}]{su2019vl}
Weijie Su, Xizhou Zhu, Yue Cao, Bin Li, Lewei Lu, Furu Wei, and Jifeng Dai.
  2020.
\newblock Vl-bert: Pre-training of generic visual-linguistic representations.
\newblock In \emph{ICLR}.

\bibitem[{Tan and Bansal(2019)}]{tan2019lxmert}
Hao Tan and Mohit Bansal. 2019.
\newblock Lxmert: Learning cross-modality encoder representations from
  transformers.
\newblock In \emph{Proceedings of the 2019 Conference on Empirical Methods in
  Natural Language Processing and the 9th International Joint Conference on
  Natural Language Processing (EMNLP-IJCNLP)}, pages 5103--5114.

\bibitem[{Vaswani et~al.(2017)Vaswani, Shazeer, Parmar, Uszkoreit, Jones,
  Gomez, Kaiser, and Polosukhin}]{vaswani2017attention}
Ashish Vaswani, Noam Shazeer, Niki Parmar, Jakob Uszkoreit, Llion Jones,
  Aidan~N Gomez, {\L}ukasz Kaiser, and Illia Polosukhin. 2017.
\newblock Attention is all you need.
\newblock In \emph{Advances in neural information processing systems}, pages
  5998--6008.

\bibitem[{Vuli{\'c} et~al.(2016)Vuli{\'c}, Kiela, Clark, and
  Moens}]{vulic2016multi}
Ivan Vuli{\'c}, Douwe Kiela, Stephen Clark, and Marie~Francine Moens. 2016.
\newblock Multi-modal representations for improved bilingual lexicon learning.
\newblock In \emph{Proceedings of the 54th Annual Meeting of the Association
  for Computational Linguistics (Volume 2: Short Papers)}, pages 188--194.

\bibitem[{Wang et~al.(2019)Wang, Singh, Michael, Hill, Levy, and
  Bowman}]{wang2018glue}
Alex Wang, Amanpreet Singh, Julian Michael, Felix Hill, Omer Levy, and Samuel
  Bowman. 2019.
\newblock Glue: A multi-task benchmark and analysis platform for natural
  language understanding.
\newblock In \emph{ICLR}.

\bibitem[{Williams et~al.(2018)Williams, Nangia, and
  Bowman}]{williams2018broad}
Adina Williams, Nikita Nangia, and Samuel Bowman. 2018.
\newblock A broad-coverage challenge corpus for sentence understanding through
  inference.
\newblock In \emph{Proceedings of the 2018 Conference of the North American
  Chapter of the Association for Computational Linguistics: Human Language
  Technologies, Volume 1 (Long Papers)}, pages 1112--1122.

\bibitem[{Wolf et~al.(2019)Wolf, Debut, Sanh, Chaumond, Delangue, Moi, Cistac,
  Rault, Louf, Funtowicz, and Brew}]{Wolf2019HuggingFacesTS}
Thomas Wolf, Lysandre Debut, Victor Sanh, Julien Chaumond, Clement Delangue,
  Anthony Moi, Pierric Cistac, Tim Rault, R'emi Louf, Morgan Funtowicz, and
  Jamie Brew. 2019.
\newblock Huggingface's transformers: State-of-the-art natural language
  processing.
\newblock \emph{ArXiv}, abs/1910.03771.

\bibitem[{Wu et~al.(2019)Wu, Ive, Wang, Madhyastha, and
  Specia}]{wu2019predicting}
Zixiu Wu, Julia Ive, Josiah Wang, Pranava Madhyastha, and Lucia Specia. 2019.
\newblock Predicting actions to help predict translations.
\newblock In \emph{ICML The How2 Challenge: New Tasks for Vision and Language
  Workshop}.

\bibitem[{Xie et~al.(2017)Xie, Girshick, Doll{\'a}r, Tu, and
  He}]{xie2017aggregated}
Saining Xie, Ross Girshick, Piotr Doll{\'a}r, Zhuowen Tu, and Kaiming He. 2017.
\newblock Aggregated residual transformations for deep neural networks.
\newblock In \emph{Proceedings of the IEEE conference on computer vision and
  pattern recognition}, pages 1492--1500.

\bibitem[{Yang et~al.(2019)Yang, Dai, Yang, Carbonell, Salakhutdinov, and
  Le}]{yang2019xlnet}
Zhilin Yang, Zihang Dai, Yiming Yang, Jaime Carbonell, Russ~R Salakhutdinov,
  and Quoc~V Le. 2019.
\newblock Xlnet: Generalized autoregressive pretraining for language
  understanding.
\newblock In \emph{Advances in neural information processing systems}, pages
  5754--5764.

\bibitem[{Young et~al.(2014)Young, Lai, Hodosh, and Hockenmaier}]{flickr30k}
Peter Young, Alice Lai, Micah Hodosh, and Julia Hockenmaier. 2014.
\newblock From image descriptions to visual denotations: New similarity metrics
  for semantic inference over event descriptions.
\newblock \emph{TACL}, 2:67--78.

\bibitem[{Zellers et~al.(2018)Zellers, Bisk, Schwartz, and
  Choi}]{zellers2018swagaf}
Rowan Zellers, Yonatan Bisk, Roy Schwartz, and Yejin Choi. 2018.
\newblock Swag: A large-scale adversarial dataset for grounded commonsense
  inference.
\newblock In \emph{Proceedings of the 2018 Conference on Empirical Methods in
  Natural Language Processing (EMNLP)}.

\bibitem[{Zhang et~al.(2020)Zhang, Chen, Wang, Utiyama, Sumita, Li, and
  Zhao}]{Zhang2020Neural}
Zhuosheng Zhang, Kehai Chen, Rui Wang, Masao Utiyama, Eiichiro Sumita, Zuchao
  Li, and Hai Zhao. 2020.
\newblock Neural machine translation with universal visual representation.
\newblock In \emph{International Conference on Learning Representations}.

\bibitem[{Zhou et~al.(2020)Zhou, Palangi, Zhang, Hu, Corso, and
  Gao}]{zhou2019unified}
Luowei Zhou, Hamid Palangi, Lei Zhang, Houdong Hu, Jason~J Corso, and Jianfeng
  Gao. 2020.
\newblock Unified vision-language pre-training for image captioning and vqa.
\newblock In \emph{AAAI}.

\bibitem[{Zhu et~al.(2015)Zhu, Kiros, Zemel, Salakhutdinov, Urtasun, Torralba,
  and Fidler}]{zhu2015aligning}
Yukun Zhu, Ryan Kiros, Rich Zemel, Ruslan Salakhutdinov, Raquel Urtasun,
  Antonio Torralba, and Sanja Fidler. 2015.
\newblock Aligning books and movies: Towards story-like visual explanations by
  watching movies and reading books.
\newblock In \emph{Proceedings of the IEEE international conference on computer
  vision}, pages 19--27.

\end{thebibliography}
